\documentclass[a4paper, 10pt, conference]{ieeeconf}       % Use this line for a4 paper

\IEEEoverridecommandlockouts                              % This command is only needed if 
\pdfoutput=1
\usepackage{cite}
\usepackage{amsmath,amssymb,amsfonts}
\usepackage{algorithm}
\usepackage{algorithmic}
\usepackage{graphicx}
\usepackage{textcomp}
\usepackage{xcolor}
\usepackage{caption}
\usepackage[switch,pagewise,columnwise]{lineno}
\usepackage{soul}
\usepackage{subcaption}
\newcommand{\Exp}{\mathop{\mathbb E}\displaylimits}

\title{\LARGE \bf
Improving Generalization of Reinforcement Learning with Minimax Distributional Soft Actor-Critic
}

\author{Yangang Ren, Jingliang Duan, Shengbo Eben Li*, Yang Guan and Qi Sun% <-this % stops a space
\thanks{This study is supported by Beijing Natural Science Foundation with JQ18010. Special thanks should be given to Toyota for partially support this study. Y. Ren and J. Duan contributed equally to this work. All correspondence should be sent to S. E. Li.}% <-this % stops a space
\thanks{All authors are with State Key
Laboratory of Automotive Safety and Energy, School of Vehicle and
Mobility, Tsinghua University, Beijing, 100084 China. Email:
{\tt\small \{ryg18, djl15, guany17\}@mails.tsinghua.edu.cn, \{lishbo, qisun\}@mail.tsinghua.edu.cn}.}%
}

\begin{document}

\maketitle
\thispagestyle{empty}
\pagestyle{empty}

%%%%%%%%%%%%%%%%%%%%%%%%%%%%%%%%%%%%%%%%%%%%%%%%%%%%%%%%%%%%%%%%%%%%%%%%%%%%%%%%
\begin{abstract}
Reinforcement learning (RL) has achieved remarkable performance in numerous sequential decision making and control tasks. However, a common problem is that learned nearly optimal policy always overfits to the training environment and may not be extended to situations never encountered during training. For practical applications, the randomness of environment usually leads to some devastating events, which should be the focus of safety-critical systems such as autonomous driving.
In this paper, we introduce the minimax formulation and distributional framework to improve the generalization ability of RL algorithms and develop the Minimax Distributional Soft Actor-Critic (Minimax DSAC) algorithm.
Minimax formulation aims to seek optimal policy considering the most severe variations from environment, in which the protagonist policy maximizes action-value function while the adversary policy tries to minimize it. Distributional framework aims to learn a state-action return distribution, from which we can model the risk of different returns explicitly, thereby formulating a risk-averse protagonist policy and a risk-seeking adversarial policy.
We implement our method on the decision-making tasks of autonomous vehicles at intersections and test the trained policy in distinct environments. Results demonstrate that our method can greatly improve the generalization ability of the protagonist agent to different environmental variations.
\end{abstract}


%%%%%%%%%%%%%%%%%%%%%%%%%%%%%%%%%%%%%%%%%%%%%%%%%%%%%%%%%%%%%%%%%%%%%%%%%%%%%%%%
\section{Introduction}
Numerous applications of deep reinforcement learning (RL) have demonstrated great performance in a range of challenging domains such as games\cite{silver2016mastering} and autonomous driving\cite{duan2020hierarchical}.
% Numerous deep reinforcement learning (RL) algorithms have appeared over the last decade\cite{mnih2015human,schaul2015prioritized,mnih2016asynchronous,lillicrap2015continuous,schulman2015trust,schulman2017proximal,haarnoja2018soft}, and their applications 
Mainstream RL algorithms focus on optimizing policy based on the performance in the training environment, without considering its universality for situations never encountered during training. Studies showed that this could reduce the generalization ability of the learned policy\cite{packer2018assessing}\cite{zhao2019investigating}. For intelligent agents, such as autonomous vehicles, we usually need them to be able to cope with multiple situations, including unknown scenarios.
% On the other hand, RL application in autonomous driving further exacerbated this issue where the environment has high stochastic. To illustrate, consider a self-driving car scenario in which we attempt to design an agent for controlling the vehicle to pass through intersections smoothly, safely, and autonomously. A straightforward idea in standard RL is that training control policy in a simulator, which is specially constructed to simulate the real environment. Unfortunately, such a trained policy is easily prone to failure as designing accurate simulators that capture intricate complexities of inte is extremely challenging. Rather than learning in simulation, another work-flow might consist of constructing a pipeline to directly learn on the hardware system itself. Apart from memory constraints, state-of-the-art reinforcement learning algorithms exhaust hundreds to millions of agent-environment interactions before acquiring successful behaviour. Of course, such high demands on sample complexities prohibit the direct application of learning algorithms on real-systems, leaving robustness to misspecified simulators a largely unresolved problem.

A straightforward technique to improve the generalization ability of RL is training on a set of random environments. 
By randomizing the dynamics of the simulation environment, the developed policies are capable of adapting to different dynamics encountered during training \cite{peng2018sim}.
Furthermore, some works have proposed that directly adding noises to state observations can provide adversarial perturbations for the training process, which can make the learned policy more insensitive to environmental variation during testing\cite{mandlekar2017adversarially,pattanaik2018robust,DBLP:journals/corr/GoodfellowSS14}.
% Furthermore, the works of \cite{mandlekar2017adversarially} and \cite{pattanaik2018robust} were proposed by directly adding noises to state observations to provide adversarial perturbations \cite{DBLP:journals/corr/GoodfellowSS14}.
However, these approaches can scarcely capture all variations from environment, as the space of dynamic parameters could be larger than the space of possible actions.
% Alternative techniques to improve generalization include risk-sensitive policy learning.
% The classic objective in RL problem is to find a policy that maximizes an expected long-run objective such as the infinite-horizon discounted or average reward. However, in many practical applications, optimizing the expected value alone is not sufficient to avoid the potential rare occurrences of large negative outcomes, and it may be necessary to include a risk measure in the optimization process, either as the objective or as a constraint\cite{prashanth2018risk,tamar2015optimizing}. Risk is related to the stochasticity of environment and with the fact that, even an
% optimal policy (in terms of expected return) may perform poorly in some cases. Therefore, some RL algorithms choose to learn risk-aware policies by considering both the mean and variance of returns, such as mean-variance trade-off method\cite{garcia2015comprehensive} and percentile optimization methods\cite{rajeswaran2016epopt}.

Alternative techniques to improve generalization include risk-sensitive policy learning.
Generally, risk is related to the stochasticity of environment and with the fact that, even an optimal policy (in terms of expected return) may perform poorly in some cases. 
Instead, risk-sensitive policy learning includes a risk measure in the optimization process, either as the objective\cite{prashanth2018risk} or as a constraint\cite{tamar2015optimizing}. This formulation not only seeks to maximize the expected reward but to optimize the risk criteria, such that the trained policy can reduce the likelihood of failure in a varying environment. In practice, the risk is always modeled as the variance of return and the most representative algorithms include mean-variance trade-off method\cite{garcia2015comprehensive} and percentile optimization method\cite{rajeswaran2016epopt}.
However, the existing methods can only model the risk by sampling discretely some trajectories from randomized environments, rather than learn the exact return distribution.

Another technique to improve generalization across different kind of environment variations is the minimax formulation. 
% To improve the generalization ability of RL across different kind of environment variations, minimax formulation, or worst case formulation has been widely adopted. 
As a pioneering work in this field, Morimoto et al. (2005) firstly combined H-infinity control with RL to learn an optimal policy, which is the prototype of most existing minimax formulation of RL algorithms\cite{morimoto2005robust}.
They formulated a differential game in which a protagonist agent tries to learn the control law by maximizing the accumulated reward while an adversary agent aims to make the worst possible destruction by minimizing the same objective. By that way, this problem was reduced to find a minimax solution of a value function. After that, Pinto et al. (2017) extended this work with deep neural network and further proposed the Robust Adversarial Reinforcement Learning (RARL) algorithm, in which the protagonist and adversary policies are trained alternatively, with one being fixed whilst the other adapts\cite{pinto2017robust}. 
Recently, Pan et al. (2019) introduced the risk-sensitive framework into RARL to prevent the rare, catastrophic events such as automotive accidents\cite{pan2019risk}.
% To that end, the risk was modeled as the variance of value functions and the protagonist policy should not only maximize expected reward, but should also select action with low variance. Conversely, the adversary policy aims to minimize the long term expected reward and select action with high variance such that the adversary can actively seek catastrophic outcomes.
% To obtain the risk of value function, they use an ensemble of Q-value networks to estimate variance, in which multiple Q-networks are trained in parallel. The risk aware RARL is effective and even crucial sometimes, especially in safety-critical systems like autonomous driving. However, the existing methods can only handle the discrete and low-dimensional action spaces and even worse, the value function must be divided into multiple discrete intervals in advance. This is inconvenient because different tasks usually require different division numbers.
For that propose, the risk was modeled as the variance of value functions and they used an ensemble of Q-value networks to estimate variance, in which multiple Q-networks were trained in parallel. Their experiments on autonomous driving demonstrated that the introduction of risk-sensitive framework into RARL is effective and even crucial, especially in safety-critical systems. 
However, the existing methods can only handle the discrete and low-dimensional action spaces, as they select actions according to their Q-networks. More urgently, the value function must be divided into multiple discrete intervals in advance. This is inconvenient because different tasks usually require different division numbers.

In this paper, we propose a new RL algorithm to improve the generalization ability of the learned policy. In particular, the learned policy can not only succeed in the training environment but also cope with the situations never encountered before. To that end, we adopt the minimax formulation, which augments the standard RL with an adversarial policy, to develop a minimax variant of Distributional Soft Actor-Critic (DSAC) algorithm\cite{duan2020addressing}, called Minimax DSAC. Here, we choose DSAC as the basis of our algorithm, not only because it is the state-of-the-art RL algorithm, but also it can directly learn a continuous distribution of returns, which enables us to model return variance as risk explicitly. By modeling risk, we can train stronger adversaries and through competition, the protagonist policy will have a greater ability to cope with environmental changes. Additionally, the application of our algorithm on autonomous driving tasks shows that Minimax DSAC can guarantee the good performance even when the environment changes drastically.

The rest of the paper is organized as follows: Section II states the preliminaries and Section III introduces formulation and implementation of the proposed method Minimax DSAC. Section IV introduces the simulation scenarios and evaluates the trained model. Section V summarizes the major contributions and concludes this paper.
\section{Preliminaries}\label{preliminaries}
Before delving into the details of our algorithm, we first introduce notation and summarize maximum entropy RL and distributional RL mathematically.
\subsection{Notation}
Standard Reinforcement Learning (RL) is designed to solve sequential decision-making tasks wherein the agent interacts with the environment.
Formally, we consider an infinite horizon discounted Markov Decision Process (MDP), defined by the tuple ($\mathcal{S}, \mathcal{A}, p, R, \gamma$), where $\mathcal{S}$ is a continuous set of states and $\mathcal{A}$ is a continuous set of actions, $p: \mathcal{S} \times \mathcal{A} \times \mathcal{S} \rightarrow \mathbb{R}$ is the transition probability distribution, $R: \mathcal{S} \times \mathcal{A} \rightarrow \mathbb{R}$ is the reward function, and $\gamma \in (0, 1]$ is the discounted factor.
In each time step $t$, the agent receives a state $s_t\in \mathcal{S}$ and selects an action $a_t \in \mathcal{A}$, and the environment will return  the next state $s_{t+1}\in \mathcal{S}$ with the probability $p(s_{t+1}|s_t,a_t)$ and a scalar reward $r_t \sim R(s_t,a_t)$.
We will use $\rho_{\pi}(s_t)$ and $\rho_{\pi}(s_t,a_t)$ to denote the state and state-action distribution induced by policy $\pi$ in environment. For the sake of simplicity, the current and next state-action pairs
are also denoted as $(s, a)$ and $(s', a')$, respectively.

\subsection{Maximum entropy RL}
The maximum entropy RL aims to maximize the expected accumulated reward and policy entropy, by augmenting the standard RL objective with an entropy maximization term:
\begin{equation}
\label{eq.policy_objective}
J_{\pi} = \mathop{\mathbb{E}}_{(s_i,a_i)\sim\rho_{\pi}}\Big[\sum^{\infty}_{i=t}\gamma^{i-t} [r(s_i,a_i)+\alpha\mathcal{H}(\pi(\cdot|s_i))]\Big],
\end{equation}
where $\alpha$ is the temperature parameter which determines the relative importance of the entropy term against the reward, and thus controls the stochasticity of the optimal policy.
The Q-value of policy $\pi$ is defined as:
\begin{equation}
\begin{aligned}
\label{eq.Q_definition}
Q^{\pi}(s_t,a_t)=\mathbb{E}[r_t]+\mathbb{E}[\sum^{\infty}_{i=t+1}\gamma^{i-t} [r(s_i,a_i)-\alpha \log\pi(a_i|s_i)]],
% Q^{\pi}(s_t,a_t)=\mathbb{E}[r_t]+\mathop{\mathbb{E}}_{(s_i,a_i)\sim\rho_{\pi}}[\sum^{\infty}_{i=t+1}\gamma^{i-t} [r(s_i,a_i)-\alpha \log\pi(a_i|s_i)]]
\end{aligned}
\end{equation}
where $r_t \sim R(s_t,a_t)$ and $(s_i, a_i) \sim \rho_\pi$.

Obviously, the maximum entropy objective differs from the maximum expected reward objective used in standard RL, though the conventional objective can be recovered as $\alpha \rightarrow 0$.
Prior works have demonstrated that the maximum entropy objective can incentive the policy to explore more widely. In problem settings where multiple actions seem equally attractive, the policy will act as randomly as possible to perform those actions\cite{haarnoja2018soft}.

% {\color{red}{The optimal maximum entropy policy is learned by a maximum entropy variant of the policy iteration method which alternates between policy evaluation and policy improvement, called soft policy iteration.}}
% In the policy evaluation process, given policy $\pi$, Q-value can be learned by repeatedly applying a modified Bellman operator $\mathcal{T^{\pi}}$ under policy $\pi$ given by
% \begin{equation}
% \nonumber
% \begin{aligned}
% \mathcal{T^{\pi}}Q^{\pi}(s,a)=&\mathbb{E}[R(s,a,s')+Q^{\pi}(s',a')-\alpha \log\pi(a'|s')].
% \end{aligned}
% \end{equation}

% The goal of the policy improvement process is to find a new policy $\pi_{\rm{new}}$ that is better than the current policy $\pi_{\rm{old}}$, such that $Q(\pi_{\rm{new}})\ge Q(\pi_{\rm{old}})$ for all state action pairs $(s,a)$.
% Hence, we can directly update the policy directly by maximizing the the entropy-augmented objective \eqref{eq.policy_objective}, i.e.,
% \begin{equation}
% \label{eq.policy_imp}
% \begin{aligned}
% \pi_{\rm{new}} &=\arg\max_{\pi} Q(\pi_{\rm{old}})\\
% &=\arg\max_{\pi} \mathbb{E}_{(s,a)\sim\rho_{\pi}}\big[Q^{\pi_{\rm{old}}}(s,a)-\alpha \log\pi(a|s)\big].
% \end{aligned}
% \end{equation}
% It has shown that policy evaluation step and policy improvement step can alternately roll forward and gradually shift to the optimal policies \cite{haarnoja2018soft}. 

\subsection{Distributional RL}
Distributional framework has attracted much attention for the reason that distributional RL algorithms show improved sample complexity and final performance.
The core idea of distributional RL is that the return
\begin{equation}
Z^{\pi}(s_t,a_t)=r(s_t,a_t)+\sum^{\infty}_{i=t+1}\gamma^{i-t} [r(s_i,a_i)-\alpha \log\pi(a_i|s_i)],
\end{equation}
is viewed as a random variable where $a_i \sim \pi(\cdot|s_i)$ and we choose to directly learn its distribution instead just its expected value, i.e., Q-value in \eqref{eq.Q_definition}:
\begin{equation}
\nonumber
Q^{\pi}(s,a)=\mathbb{E}[Z^{\pi}(s,a)].
\end{equation}
Under this theme, many works used discrete distribution to build the return distribution, in which we need to divide the value function into different intervals priorly.
Recently, Duan et al.\cite{duan2020addressing} proposed the Distributional Soft Actor-Critic (DSAC) algorithm  to directly learn the continuous distribution of returns by truncating the difference between the target and current return distribution.
Therefore, we draw on the continuous return distribution in the following illustration.

The optimal policy is learned by a distributional variant of the policy iteration method which alternates between policy evaluation and policy improvement.
The corresponding variant of Bellman operator can be derived as:
\begin{equation}
\nonumber
\mathcal{T^{\pi}}Z^\pi(s,a) \overset{D}{=}r(s,a)+\gamma( Z^\pi(s',a')-\log\pi(a'|s')),
\end{equation}
where $A \overset{D}{=} B$ denotes that two random variables $A$ and $B$ have equal probability laws and the next state $s'$ and action $a'$ are distributed according to $p(\cdot|s,a)$ and $\pi(\cdot|s')$ respectively.

Supposing $\mathcal{T^{\pi}}Z(s,a)\sim\mathcal{T}^{\pi}_{\mathcal{D}}\mathcal{Z}(\cdot|s,a)$, where $\mathcal{T}^{\pi}_{\mathcal{D}}\mathcal{Z}(\cdot|s,a)$ denotes the distribution of $\mathcal{T^{\pi}}Z(s,a)$, the return distribution can be optimized by minimizing the distribution distance between Bellman updated and the current return distribution: 
\begin{equation}
\label{eq.policy_eva}
\mathcal{Z}_{\rm{new}} =  \arg\min_{\mathcal{Z}}\mathop{\mathbb{E}}_{(s,a)\sim\rho_{\pi}}\big[d(\mathcal{T}^{\pi}_{\mathcal{D}}\mathcal{Z}_{\rm{old}}(\cdot|s,a),\mathcal{Z}(\cdot|s,a))\big],
\end{equation}
where $d$ is some metric to measure the distance between two distribution. For example, we can adopt $d$ as the Kullback-Leibler (KL) divergence or Wasserstein metric.
In policy improvement process, we aim to find a new policy $\pi_{\rm{new}}$ that is better than the current policy $\pi$, such that $J_{\pi_{\rm{new}}}\ge J_{\pi}$ for all state action pairs $(s,a)$:
\begin{equation}
\label{eq.policy_imp}
\begin{aligned}
\pi_{\rm{new}} &=\arg\max_{\pi} J_{\pi}\\
&=\arg\max_{\pi} \mathbb{E}_{(s,a)\sim\rho_{\pi}}\big[Q^{\pi}(s,a)-\alpha \log\pi(a|s)\big] .
\end{aligned}
\end{equation}
It has shown that policy evaluation step in \eqref{eq.policy_eva} and policy improvement step in \eqref{eq.policy_imp} can alternately roll forward and gradually shift to the optimal policies \cite{haarnoja2018soft,duan2020addressing}.
% Hence, we can directly update the policy directly by maximizing the the entropy-augmented objective \eqref{eq.policy_objective}, i.e.,
% \begin{equation}
% \label{eq.policy_imp}
% \begin{aligned}
% \pi_{\rm{new}} &=\arg\max_{\pi} Q(\pi_{\rm{old}})\\
% &=\arg\max_{\pi} \mathbb{E}_{(s,a)\sim\rho_{\pi}}\big[Q^{\pi_{\rm{old}}}(s,a)-\alpha \log\pi(a|s)\big].
% \end{aligned}
% \end{equation}
% It has shown that policy evaluation step and policy improvement step can alternately roll forward and gradually shift to the optimal policies \cite{haarnoja2018soft}. 
% In the policy evaluation process, given policy $\pi$, Q-value can be learned by repeatedly applying a modified Bellman operator
% \begin{equation}
% \nonumber
% \mathcal{T^{\pi}}Z^\pi(s,a) \overset{D}{=}r(s,a)+\gamma( Z^\pi(s',a')-\log\pi(a'|s'))
% \end{equation}
% can be derived, 

% However, many prior works used discrete distribution to build the return distribution, in which we need to divide the value function into different intervals priorly.
\section{Our methods}
% This section mainly focuses on the combination of minimax formulation and distributional RL framework, in which we state our algorithm based on the continuous distributional return.
Although distributional RL algorithms like DSAC considered the randomness of return caused by the environment, they may still fail in a distinct environment. Here, we introduce the minimax formulation into the existing DSAC algorithm and model the risk explicitly through the continuous return distribution.

\subsection{Minimax Distributional Soft Actor-Critic (Minimax DSAC)}
In minimax formulation, there exist two policies to be optimized, called protagonist policy and adversary policy respectively. 
Given the current state $s_t$, the protagonist policy $\pi_a$ will take action $a_t \in \mathcal{A}$, the adversary policy $\pi_u$ will take action $u_t \in\mathcal{U}$, and then the next state $s_{t+1}$ will be reached. Whereas these two policies obtain different rewards: the protagonist gets a reward $r_t$ while the adversary gets a reward $-r_t$ at each time step.
We use $\rho_{\pi}(s_t)$ and $\rho_{\pi}(s_t,a_t,u_t)$ to denote the state and state-action distribution induced by policy $\pi_a$ and $\pi_u$ in environment.

Under this theme, the random return generated by $\pi_a$ and $\pi_u$ can be rewritten as:
\begin{equation}
\nonumber
\begin{aligned}
&Z(s_t,a_t, u_t):=Z^{\pi_a,\pi_u}(s_t,a_t, u_t)
\\
&=r(s_t,a_t,u_t)+\sum^{\infty}_{i=t+1}\gamma^{i-t} [r(s_i,a_i,u_i)-\alpha \log\pi_a(a_i|s_i)],
\end{aligned}
\end{equation}
and its expectation is the action value function $Q$:
\begin{equation}
\nonumber
Q(s,a,u)=\mathbb{E}\left[Z(s,a,u)\right].
\end{equation}
% For the sake of brevity, we will denote $Q^{\pi_a,\pi_u}(s,a,u), Z^{\pi_a,\pi_u}(s,a,u)$ as $Q(s,a,u), Z(s,a,u)$ respectively and 
Suppose $Z(s,a,u)\sim \mathcal{Z}(s,a,u)$, we can use the similar method in \eqref{eq.policy_eva} to update the return distribution. To learn risk-sensitive policies, we model risk as the variance of the learned continuous return distribution, where the protagonist policy is optimized to mitigate risk to avoid the potential events that have the chance to lead to bad return, i.e., maximizing the following objective:
% The corresponding minimax distributional Bellman operator can derived as:
% \begin{equation}
% Z(s,a,u) \overset{D}{=}r(s,a,u,s')+\gamma( Z(s',a',u')),
% \end{equation}
% where $s'\sim p(\cdot|s,a,u)$, $a'\sim \pi_a(\cdot|s')$ and $u'\sim \pi_u(\cdot|s')$.
%In policy improvement step, both protagonist policy and adversary policy optimize themselves based on current return distribution, in which they have common objective function:
% The protagonist policy aims to maximize the distributional expected return while the adversary aims to minimize it:
% \begin{equation}\label{minimax equation}
% \min_{\pi_{u}}\max_{{\pi_{a}}}
% J(\pi_{a}, \pi_{u} ) .
% \end{equation}
\begin{equation}
\begin{aligned}
J(\pi_{a})
=\mathbb{E}_{(s,a,u)\sim\rho_{\pi}}&\Big[Q(s,a,u)-\lambda_a \sigma(s,a,u)\Big].
\end{aligned}
\label{eq:7}
\end{equation}
And the adversary policy seeks to increase risk to disrupt the learning process, i.e., minimizing the following objective:
\begin{equation}
\begin{aligned}
J(\pi_{u})
=\mathbb{E}_{(s,a,u)\sim\rho_{\pi}}&\Big[Q(s,a,u)-\lambda_u \sigma(s,a,u)\Big],
\end{aligned}
\label{eq:8}
\end{equation}
where $\lambda_a \ge 0$ and $\lambda_u \ge 0$ are the constants corresponding to the variance $\sigma(s,a,u)$ which describes different risk level.
\subsection{Implementation of Minimax DSAC}
To handle problems with large continuous domains, we use function approximators for all the return distribution function and two policies, which can be modeled as a Gaussian with the mean and variance given by neural networks (NNs). 
We will consider a parameterized state-action return distribution function $\mathcal{Z}_{\theta}(s,a,u)$, a stochastic protagonist policy $\pi_{\phi}(a|s)$ and a stochastic adversarial policy $\pi_{\mu}(u|s)$ where $\theta$, $\phi$ and $\mu$ are parameters. Next we will derive update rules for these parameter vectors and show the details of our Minimax DSAC.

In policy evaluation step, the return distribution is updated by minimizing the difference between the target return distribution and the current return distribution. The formulation is similar with the DSAC algorithm  except that we consider two policies\cite{duan2020addressing}:
\begin{equation}
\nonumber
\begin{aligned}
&J_{\mathcal{Z}}(\theta) \\
&=  \mathop{\mathbb{E}}_{(s,a,u)\sim\rho_{\pi}}\big[D_{\rm{KL}}(\mathcal{T}^{\pi_{\phi'},\pi_{\mu'}}_{\mathcal{D}}\mathcal{Z}_{\theta'}(\cdot|s,a,u),\mathcal{Z}_{\theta}(\cdot|s,a,u))\big]\\
&={\rm{c}}- \Exp_{\substack{(s,a,u) \sim\rho_{\pi}, \\Z(s',a',u')\sim\mathcal{Z}_{\theta'}}}\Big[\log\mathcal{P}(\mathcal{T}^{\pi_{\phi'},\pi_{\mu'}}_{\mathcal{D}} Z(s,a,u)|\mathcal{Z}_{\theta})\Big],
\end{aligned}
\end{equation}
where $c$ is a constant. The gradient about parameter $\theta$ can be written as:
\begin{equation}
\nonumber
\begin{aligned}
&\nabla_{\theta}J_{\mathcal{Z}}(\theta) \\
&=   -\Exp_{\substack{(s,a,u) \sim\rho_{\pi}, \\Z(s',a',u')\sim\mathcal{Z}_{\theta'}}}\Big[\nabla_{\theta}\log\mathcal{P}(\mathcal{T}^{\pi_{\phi'},\pi_{\mu'}}_{\mathcal{D}}Z(s,a,u)|\mathcal{Z}_{\theta})\Big].
\end{aligned}
\end{equation}
To prevent the gradient exploding, we adopt the clipping technique to keep it close to the expectation value $Q_{\theta}(s,a,u)$ of the current distribution $\mathcal{Z}_{\theta}(s,a,u)$:
\begin{equation}
\nonumber
\begin{aligned}
&\mathcal{T}^{\pi_{\phi'},\pi_{\mu'}}_{\mathcal{D}}Z(s,{a},{u})\\
&={\rm{clip}}(\mathcal{T}^{\pi_{\phi'},\pi_{\mu'}}_{\mathcal{D}}Z(s,{a},{u}),Q_{\theta}(s,{a},{u})-b,Q_{\theta}(s,{a},{u})+b),
\end{aligned}
\end{equation}
where $b$ is a hyperparameter representing the clipping boundary.
To stabilize the learning process, target return distribution with parameter $\theta'$, two policy functions with separate parameters $\phi'$ and $\mu'$, are used to evaluate the target function. The target networks use a slow-moving update rate, parameterized by $\tau$, such as 
% \begin{equation}
% \label{eq.target_update}
% \begin{aligned}
% \theta' \leftarrow  \tau\theta+(1-\tau)\theta',\\
% \phi' \leftarrow  \tau\phi+(1-\tau)\phi', \\
% \mu' \leftarrow  \tau\mu+(1-\tau)\mu'. 
% \end{aligned}
% \end{equation}
\begin{equation}
\label{eq.target_update}
\begin{aligned}
x' \leftarrow  \tau x+(1-\tau)x',
\end{aligned}
\end{equation}
where $x$ represents the parameters $\theta$, $\mu$ and $\phi$.

In policy improvement step, as discussed in \eqref{eq:7}, the protagonist policy aims to maximize the expected return with entropy and select actions with low variance:
\begin{equation}
\begin{aligned}
J(\phi)=\mathbb{E}\Big[\mathbb{E}_{Z(s,a,u)
\sim\mathcal{Z}_{\theta}}[Z(s,a,u)]-\lambda_a \sigma(Z(s,a,u))\Big].
\end{aligned}
\label{equ:pro}
\end{equation}
The adversarial policy in \eqref{eq:8} aims to minimize the expected return and select actions with high variance:
\begin{equation}
\begin{aligned}
J(\mu)=\mathbb{E}\Big[\mathbb{E}_{Z(s,a,u)
\sim\mathcal{Z}_{\theta}}[Z(s,a,u)]-\lambda_u \sigma(Z(s,a,u))\Big].
\end{aligned}
\label{equ:adv}
\end{equation}
Suppose the mean $Q_\theta$ and variance $\sigma_\theta$ of the return distribution can be explicitly parameterized by parameters $\theta$.
We can derive the policy gradient of protagonist and adversary policy using the reparameterization trick:
\begin{equation}
\nonumber
a=f_{\phi}(\xi_{a};s),\qquad
u=h_{\mu}(\xi_{u};s),
\end{equation}
% \begin{equation}
% \nonumber
% u=h_{\mu}(\xi_{u};s),
% \end{equation}
where $\xi_{a}$, $\xi_{u}$ is auxiliary variables which are sampled from some fixed distribution. Then the protagonist policy gradient of \eqref{equ:pro} can be derived as:
\begin{equation}
\nonumber
\begin{aligned}
\partial_{\phi}J(\phi)=\mathbb{E}_{\rho_{\pi},\xi_a}\Big[&-\nabla_{\phi}\alpha\log(\pi_{\phi}(a|s))+
(\nabla_{a}Q_{\theta}(s,a,u)\\
&-\alpha\nabla_{a}\log(\pi_{\phi}(a|s)))\nabla_{\phi}f_{\phi}(\xi_{a};s)\\
&-\lambda_a\nabla_{a}\sigma_\theta(s,a,u)\nabla_{\phi}f_{\phi}(\xi_{a};s)\Big].
\end{aligned}
\end{equation}
And the adversarial policy gradient of \eqref{equ:adv} can be approximated with
\begin{equation}
\nonumber
\begin{aligned}
\partial_{\mu}J(\mu)
=\mathbb{E}_{\rho_{\pi},\xi_u}\Big[\nabla_{u}Q_{\theta}&(s,a,u)\nabla_{\mu}h_{\mu}(\xi_{u};s)\\
&- \lambda_u\nabla_{u}\sigma_\theta (s,a,u)\nabla_{\mu}h_{\mu}(\xi_{\mu};s)\Big].
\end{aligned}
\end{equation}
Finally, the temperature $\alpha$ is updated by minimizing the following objective
\begin{equation}
\nonumber
J(\alpha)=\mathbb{E}_{(s,a,u)\sim\rho_{\pi}}[-\alpha \log\pi_{\phi}(a|s)-\alpha\overline{\mathcal{H}}],
\end{equation}
where $\overline{\mathcal{H}}$ is the expected entropy.
The detail of our algorithm can be shown as Algorithm \ref{alg:Minimax DSAC}.
\begin{algorithm}[!htb]
\caption{Minimax DSAC Algorithm}
\nonumber
\label{alg:Minimax DSAC}
\begin{algorithmic}
\STATE Initialize parameters $\theta$, $\phi$, $\mu$ and $\alpha$
\STATE Initialize target parameters $\theta'\leftarrow\theta$, $\phi'\leftarrow\phi$, $\mu'\leftarrow\mu$
\STATE Initialize learning rate $\beta_{\mathcal{Z}}$, $\beta_{\phi}$, $\beta_{\mu}$, $\beta_{\alpha}$ and $\tau$ 
\REPEAT
\STATE Select action $a^1\sim\pi_{\phi}(a|s)$, $a^2\sim\pi_{\mu}(a|s)$
\STATE Observe reward $r$ and new state $s'$
\STATE Store transition tuple $(s,a^1, a^2, r,s')$ in buffer $\mathcal{B}$
\STATE
\STATE Sample $N$ transitions $(s,a^1, a^2, r,s')$ from $\mathcal{B}$
\STATE Update return distribution $\theta \leftarrow \theta - \beta_{\mathcal{Z}}\nabla_{\theta}J_{\mathcal{Z}}(\theta)$
% \IF{$k$ mod $m$}
\STATE Update protagonist policy $\phi \leftarrow \phi + \beta_{\phi}\partial_{\phi} J(\phi)$
\STATE Update adversarial policy $\mu \leftarrow \mu - \beta_{\mu}\partial_{\mu} J(\mu)$

\STATE Adjust temperature $\alpha \leftarrow \alpha - \beta_{\alpha}\nabla_{\alpha} J(\alpha)$
\STATE Update target networks using \eqref{eq.target_update}
\UNTIL Convergence  
\end{algorithmic}
\end{algorithm}
\section{Experiments}
In this section, we evaluate our algorithm on an autonomous driving tasks, in which we choose the intersection as the driving scenario. 
% Our experiment aims to study two primary questions: (1) How well does Minimax DSAC perform on this task in terms  standard  DSAC algorithm? (2) Can our algorithm still work or behave better if there are some variations from environment?
\subsection{Simulation Environment}
\begin{figure}[thbp]
% \captionsetup{justification =centering,
%               singlelinecheck = false,labelsep=period, font=small}
\centering{\includegraphics[width=0.6\linewidth]{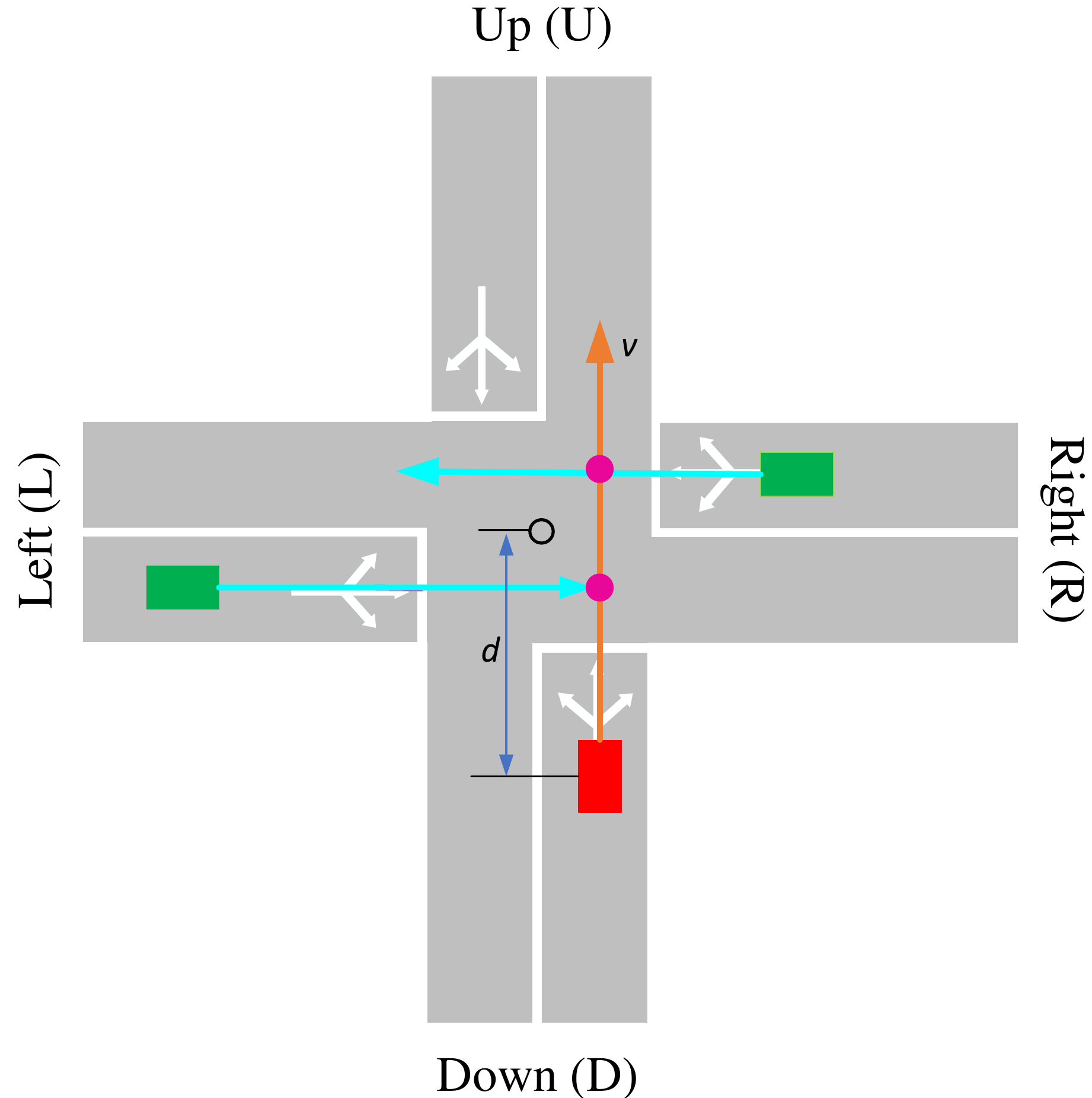}}
\caption{Intersection Scenario.}
\label{intersection}
\end{figure}

We focus on a typical 4-direction intersection shown in Fig.~\ref{intersection}. Each direction is denoted by its location, i.e. up(U), down (D), left (L) and right (R) respectively.
The intersection is unsignalized and each direction has one lane.
The protagonist vehicle (red car in Fig.~\ref{intersection}) attempts to travel from down to up, while two adversarial vehicles (green cars in Fig.~\ref{intersection}) ride from right to left, left to right respectively. The trajectories of all three vehicles are given priorly, and as a result, there are two traffic conflict points in the path of protagonist vehicle and adversarial vehicles, as the solid circle shown in Fig.~\ref{intersection}. In our experiment setting, the protagonist vehicle attempts to pass the intersection safely and quickly, while the other two adversarial vehicles try to provide disruption by hitting the protagonist vehicle.

We choose position and velocity information of each vehicle as states, i.e., ($d$, $v$), where $d$ is distance between vehicle and center of the intersection. Note that $d$ is positive when a vehicle is heading for the center and negative when it is leaving. For action space, we choose the acceleration of each vehicle and suppose that vehicles can strictly follow the desired acceleration. In total, 6-dimensional continuous state space and 3-dimensional continuous action space are constructed.

The reward function is designed to consider both safety and time efficiency. This task is constructed in an episodic manner, where two terminate conditions are given: collision or passing.
First, if the protagonist vehicle passes the intersection safely, a large positive reward 110 is given; Second, if a collision happens anywhere, a large negative reward -110 is given to the protagonist vehicle; Besides, a minor negative reward -1 is given every time step to encourage the protagonist vehicle to pass as quickly as possible. However, the adversarial vehicles obtain opposite reward in every case aforementioned. 

\subsection{Algorithm Details and Results}
\begin{figure}[thbp]
\centering{\includegraphics[width=0.7\linewidth]{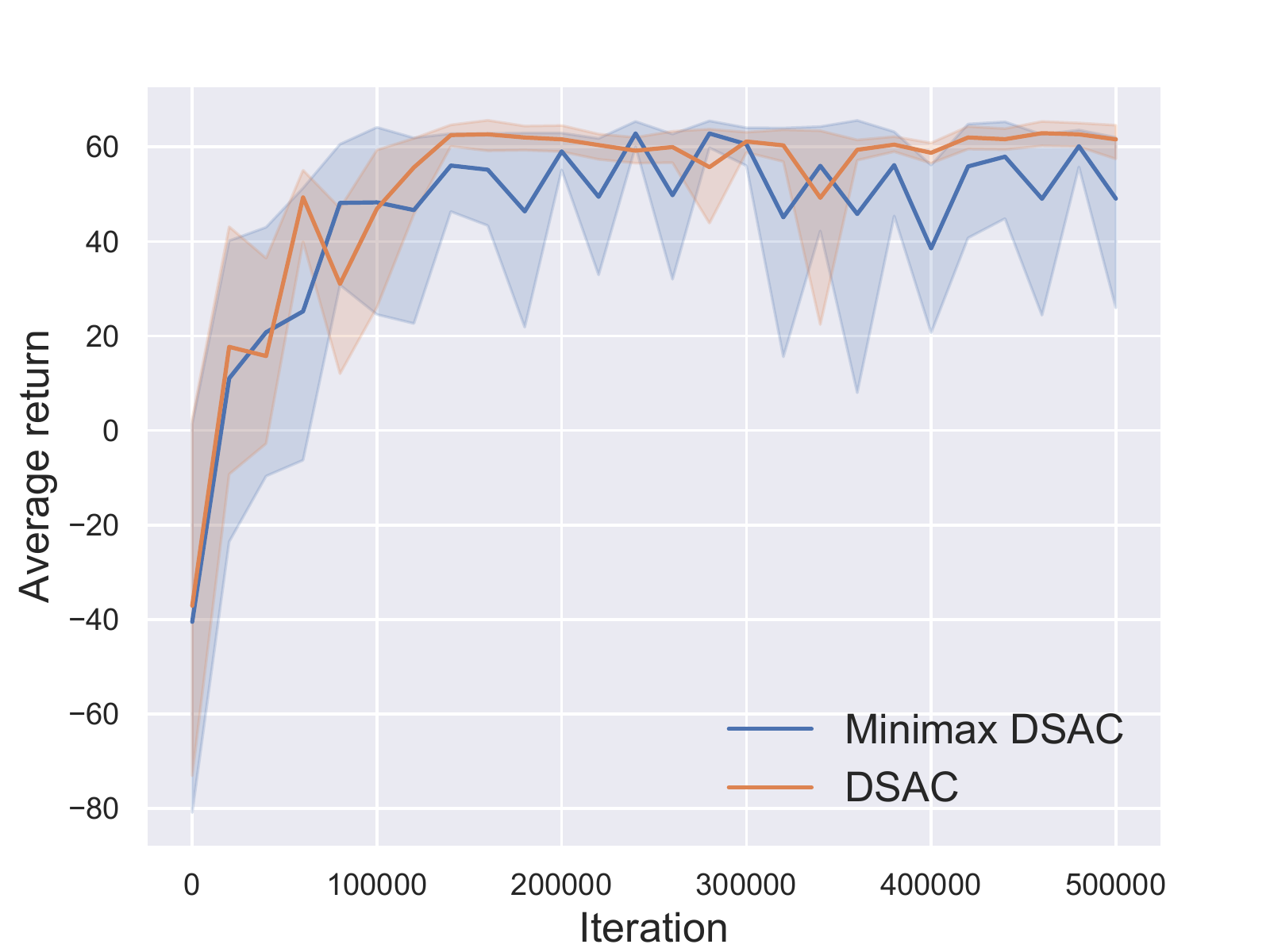}}
\caption{Average return during training process. The solid lines correspond to the mean and the shaded regions correspond to 95\% confidence interval over 10 runs.}
\label{figure:average return}

\end{figure}

Both the value function and two policies are approximated by multi-layer perceptron (MLP) with 2 hidden layers and 256 units per layer. The policy of protagonist vehicle aims to maximize future expected return, while the policy of adversarial vehicles aims to minimize it.
The baseline of our algorithm is the standard DSAC \cite{duan2020addressing} without the adversarial policy, in which the protagonist vehicle learns to pass through the intersection with the existence of two random surrounding vehicles. Also, we adopt the asynchronous parallel architecture of DSAC called PABLE, in which 4 learners and 3 actors are designed to accelerate the learning speed.
The hyperparameters used in training are listed in Table \ref{table.hyper} and the training result is shown as Fig.~\ref{figure:average return}.
\begin{table}[ht]
\captionsetup{justification=centering,labelsep=newline,font=small}
\caption{Trainning hyperparameters}
\label{table.hyper}
\vskip 0.15in
\begin{center}
\begin{small}
\begin{sc}
\begin{tabular}{c|ccc}
\hline

% Hyperparameters & DSAC & Minimax DSAC & under \\\hline
% Model type &\multicolumn{3}{c}{MLP} \\\hline
% Hidden units  & \multicolumn{3}{c}{256} \\\hline
% Hidden layers & \multicolumn{3}{c}{2}\\\hline
Max buffer size &  \multicolumn{3}{c}{500}\\\hline
Sample batch size &  \multicolumn{3}{c}{256}\\\hline
Hidden layers activation&  \multicolumn{3}{c}{gelu}\\\hline
Optimizer type &  \multicolumn{3}{c}{Adam}\\\hline
Adam parameter &\multicolumn{3}{c}{$\beta_{1}=0.9, \beta_{2}=0.999$}\\\hline
Actor learning rate & \multicolumn{3}{c}{$5{\rm{e-}}5\rightarrow5{\rm{e-}}6 $}\\\hline
Critic learning rate & \multicolumn{3}{c}{$1{\rm{e-}}4\rightarrow1{\rm{e-}}5 $}\\\hline
$\alpha$ learning rate & \multicolumn{3}{c}{$5{\rm{e-}}5\rightarrow5{\rm{e-}}6 $}\\\hline
Discount factor$\gamma$ & \multicolumn{3}{c}{0.99}\\\hline
Temperature $\alpha$ & \multicolumn{3}{c}{$5\rm{e-}5\rightarrow{5\rm{e-}}6 $}\\\hline
Target update rate $\tau$ & \multicolumn{3}{c}{0.001}\\\hline
Expected entropy $\overline{\mathcal{H}}$  & \multicolumn{3}{c}{- ACTION DIMENSIONS}\\\hline
Clipping boundary $b$ &\multicolumn{3}{c}{20}\\\hline
% Actor number&\multicolumn{3}{c}{4}\\\hline
% Learner number &\multicolumn{3}{c}{3}\\\hline
% Buffer number &\multicolumn{3}{c}{1}\\ \hline
% Seed & \multicolumn{3}{c}{Current time}\\\hline
$\lambda_a,\lambda_u $ & \multicolumn{3}{c}{0.1}\\

\hline
\end{tabular}
\end{sc}{}
\end{small}
\end{center}
\vskip -0.1in
\end{table}

Results show that Minimax DSAC obtained a smaller mean with respect to the average return, which is explicable that the adversary policy provides a strong disruption to the learning of protagonist policy. Besides, it is clear that Minimax DSAC has more fluctuation than standard DSAC at convergence stage. 
That can be explained that the protagonist vehicle has learned to avoid the potential collision by decelerating and even stopping and waiting in face of the despiteful adversarial vehicles, which will lead to punishment in each step and finally result in a lower return. 

\subsection{Evaluation}
Compared with the performance during training process, we concern more about that on situations distinct from the training environment, i.e., the generalization ability. As adversarial vehicles can be regarded as part of the environment, we can design different driving modes of adversarial agents to adjust the environment difficulty to evaluate the generalization ability of the protagonist policy. Formally, we design three driving modes for the adversarial agents: aggressive, conservative and random.
In aggressive mode, the two adversarial vehicles sample their acceleration from positive interval $[1.0, 2.0](\text{m/s}^2)$ while in conservative mode they sample acceleration from negative interval $[-2.0, -1.0](\text{m/s}^2)$.
In random mode, one adversarial vehicle samples acceleration from $[-2.0, -1.0](\text{m/s}^2)$ and the other vehicle samples acceleration from $[1.0, 2.0](\text{m/s}^2)$.

\begin{figure}[thbp]
% \captionsetup{justification =centering,
%               singlelinecheck = false,labelsep=period, font=small}
\centering{\includegraphics[width=0.7\linewidth]{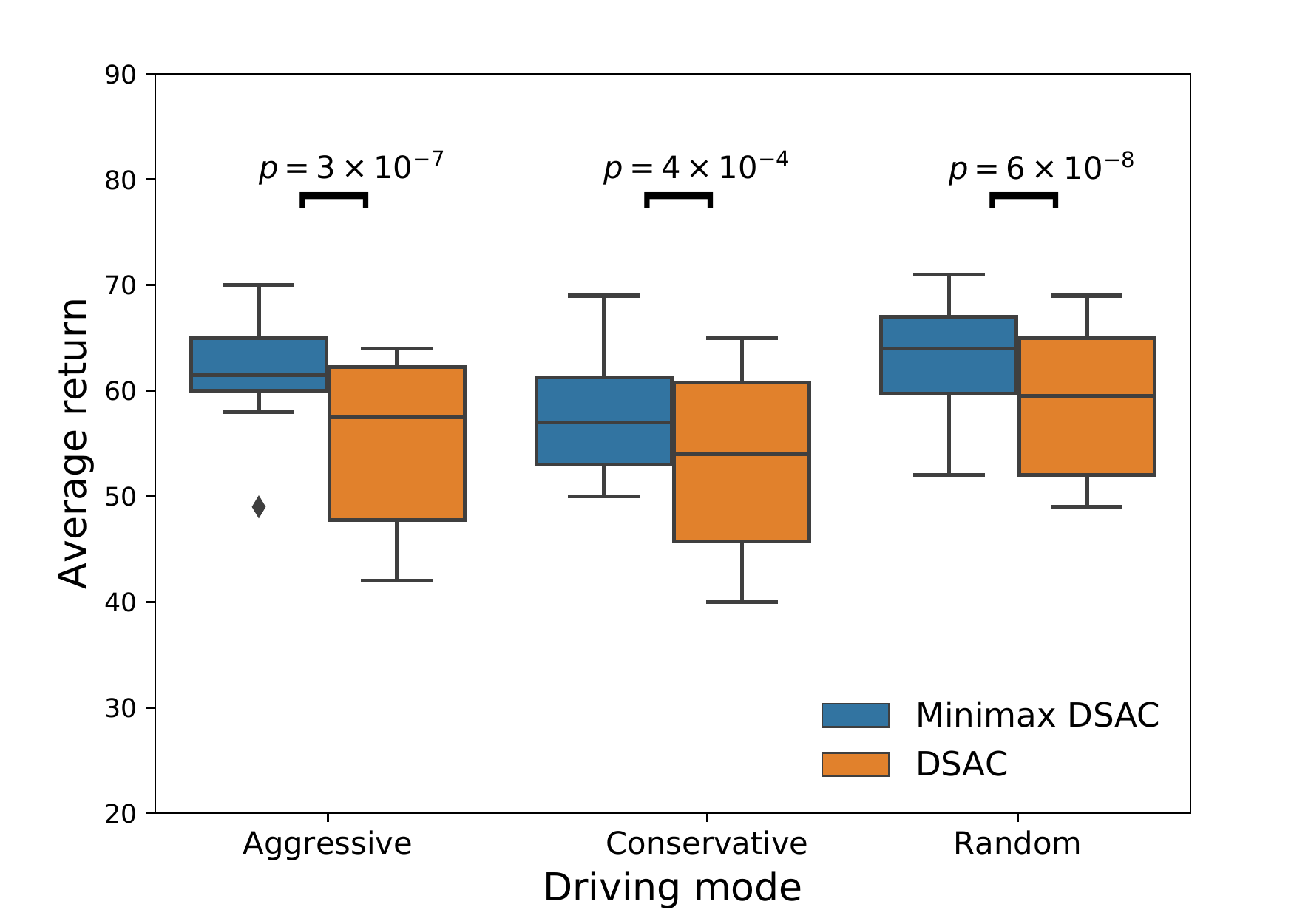}}
\caption{Average return during testing process. Each boxplot is drawn based on values of 20-episode evaluations.}
\label{figure:time}
\end{figure}

The comparison of two methods under three modes is shown in Fig.~\ref{figure:time}, in which the corresponding $p$-values are also marked. Results show that Minimax DSAC can greatly improve the performance under different modes of adversarial vehicles, especially in aggressive and random mode.
In conservative mode, these two algorithms show minor difference because both the adversarial vehicles drive at the lowest speed in the limit, thereby less potential collision to the protagonist will happen. However, Minimax DSAC still obtained a higher return because it adopted large acceleration to improve the passing efficiency.
The t-test results in Fig.~\ref{figure:time} show that the average reward of DSAC is significantly smaller than that of Minimax DSAC $(p < 0.001)$.

\begin{figure*}[thbp]
\centering
\noindent\makebox[\textwidth][c]{
\begin{subfigure}[]{0.33\textwidth}
  \centering
  \includegraphics[width=4.2cm]{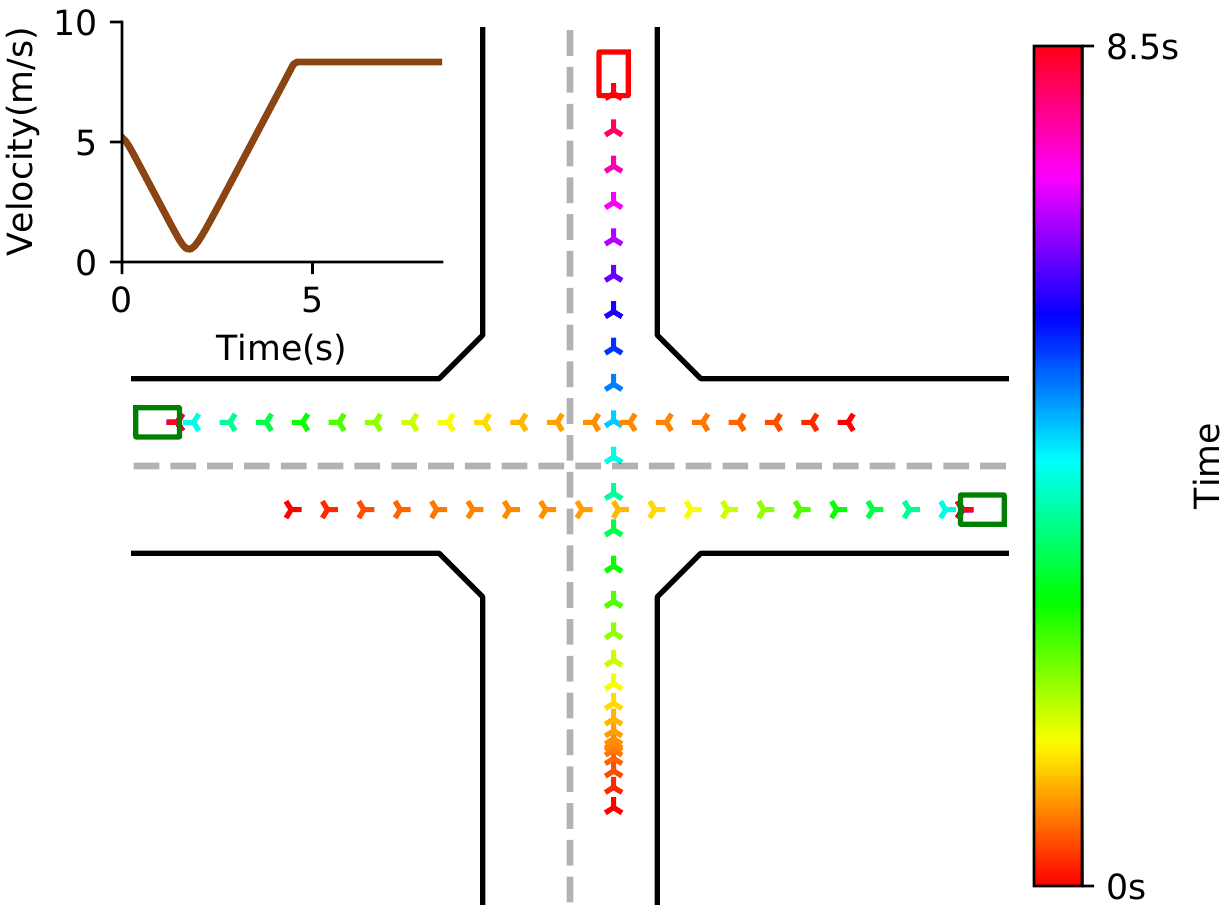}
  \caption{}
  \label{fig:agg}
\end{subfigure}
\begin{subfigure}[]{0.33\textwidth}
  \centering
  \includegraphics[width=4.2cm]{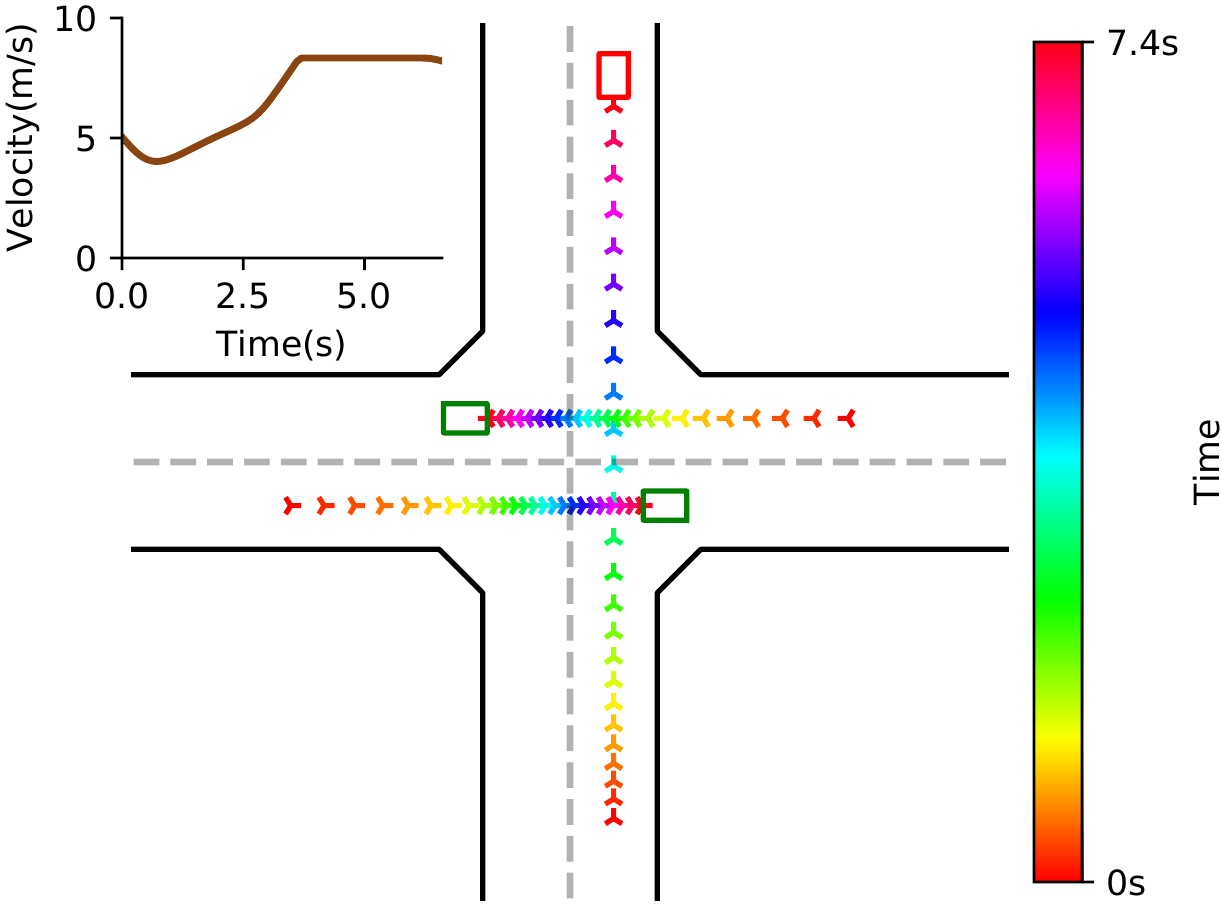}
  \caption{}
  \label{fig:Ant_ave}
\end{subfigure}
\begin{subfigure}[]{0.33\textwidth}
  \centering
  \includegraphics[width=4.2cm]{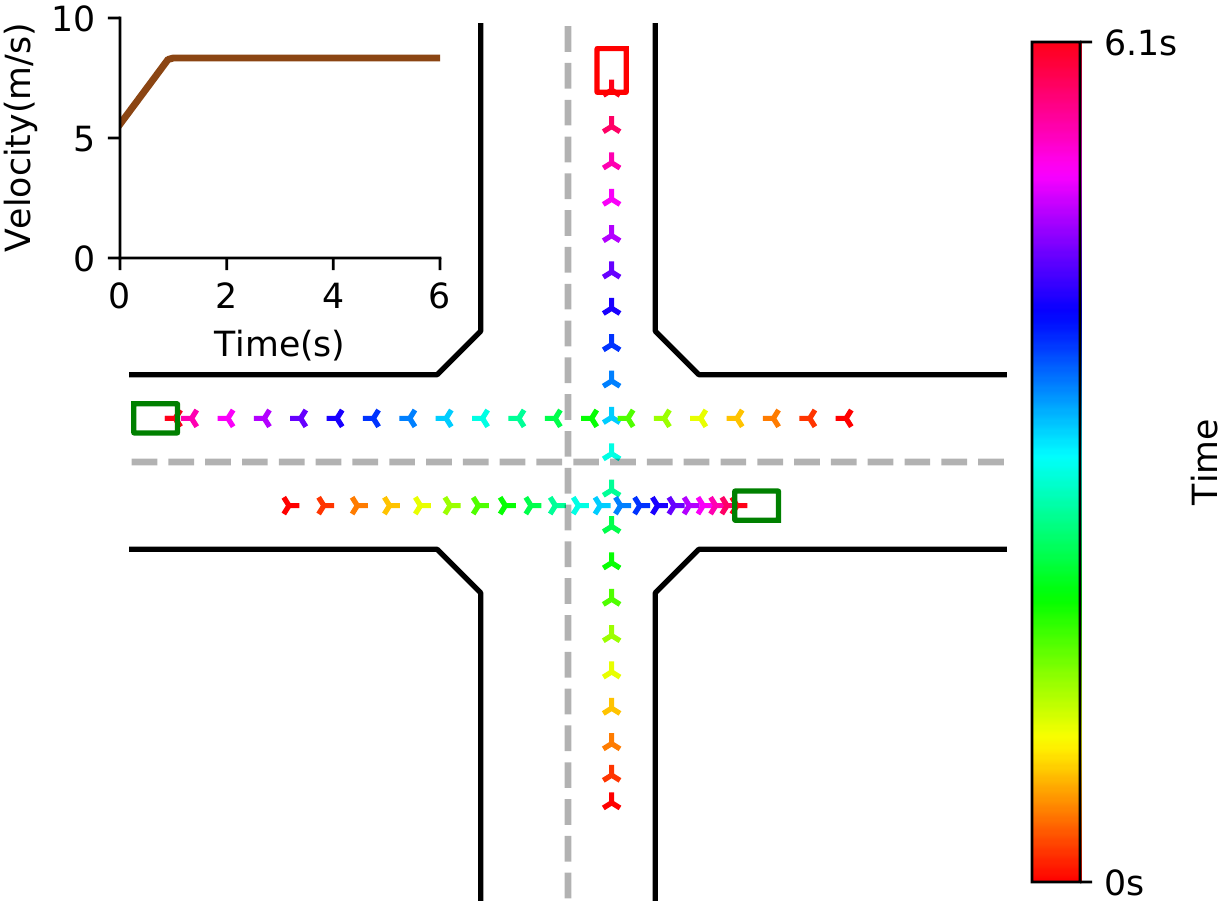}
  \caption{}
  \label{fig:HalfCheetah_ave}
\end{subfigure}
}
\noindent\makebox[\textwidth][c]{
\begin{subfigure}[]{0.33\textwidth}
  \centering
  \includegraphics[width=4.2cm]{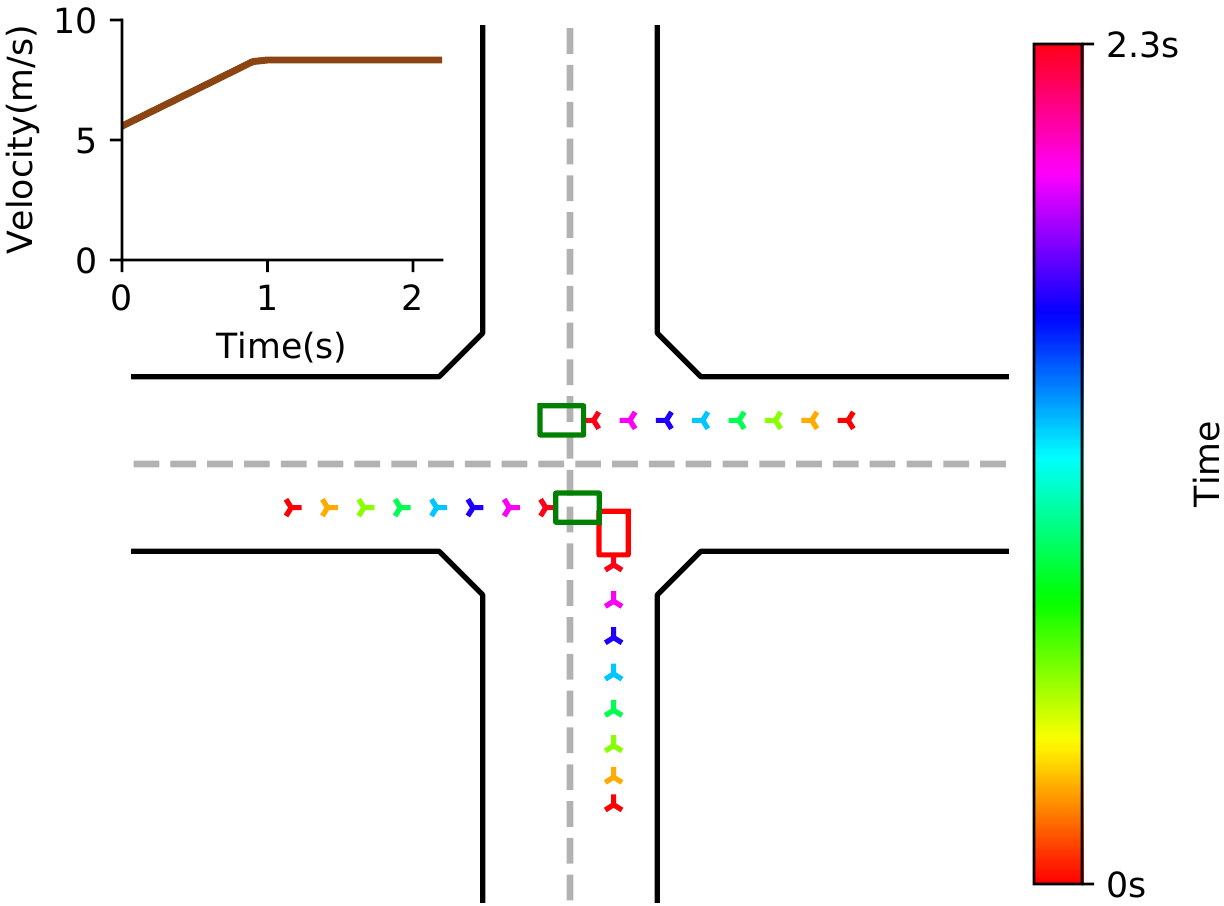}
  \caption{}
  \label{fig:InvertedDoublePendulum_ave}
\end{subfigure}
\begin{subfigure}[]{0.33\textwidth}
  \centering
  \includegraphics[width=4.2cm]{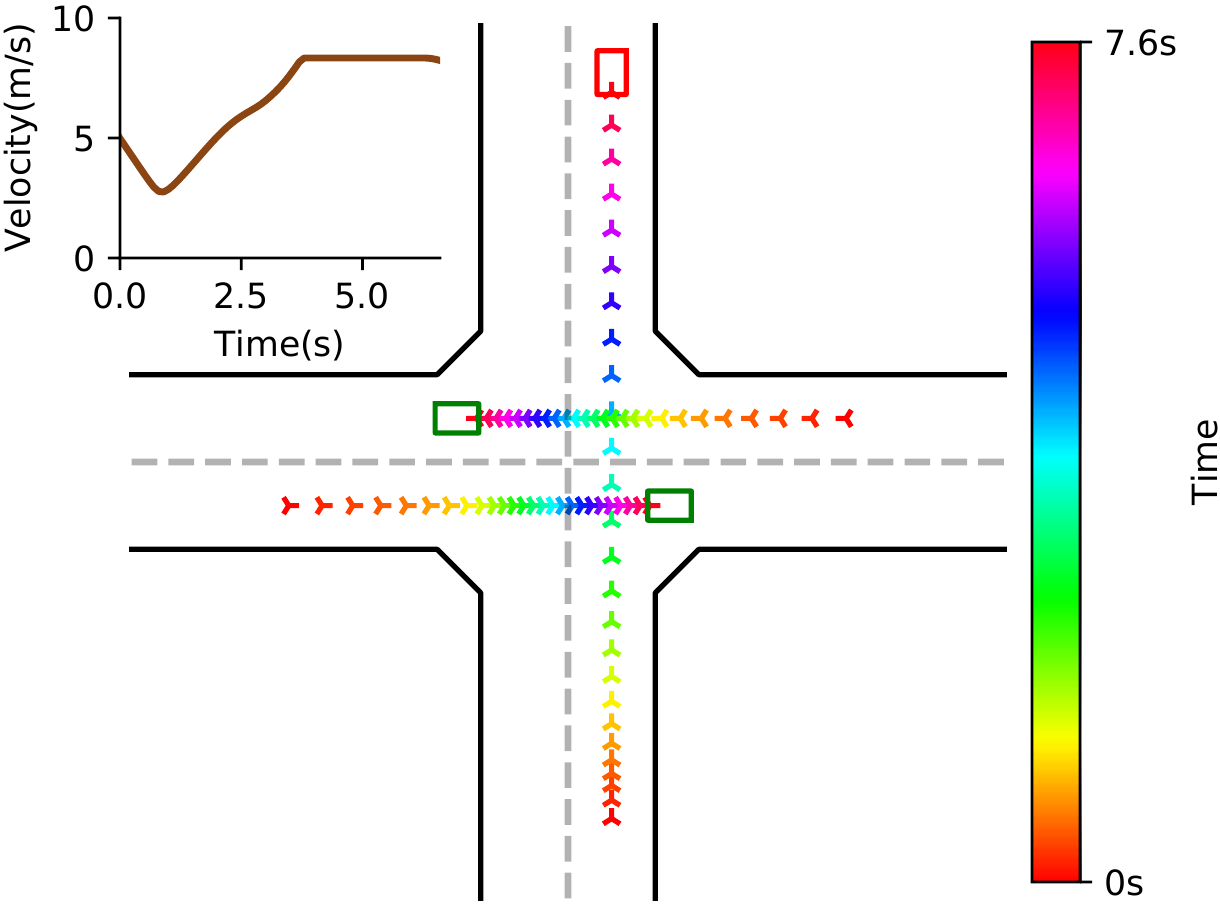}
  \caption{}
  \label{fig:Ant_ave}
\end{subfigure}
\begin{subfigure}[]{0.33\textwidth}
  \centering
  \includegraphics[width=4.2cm]{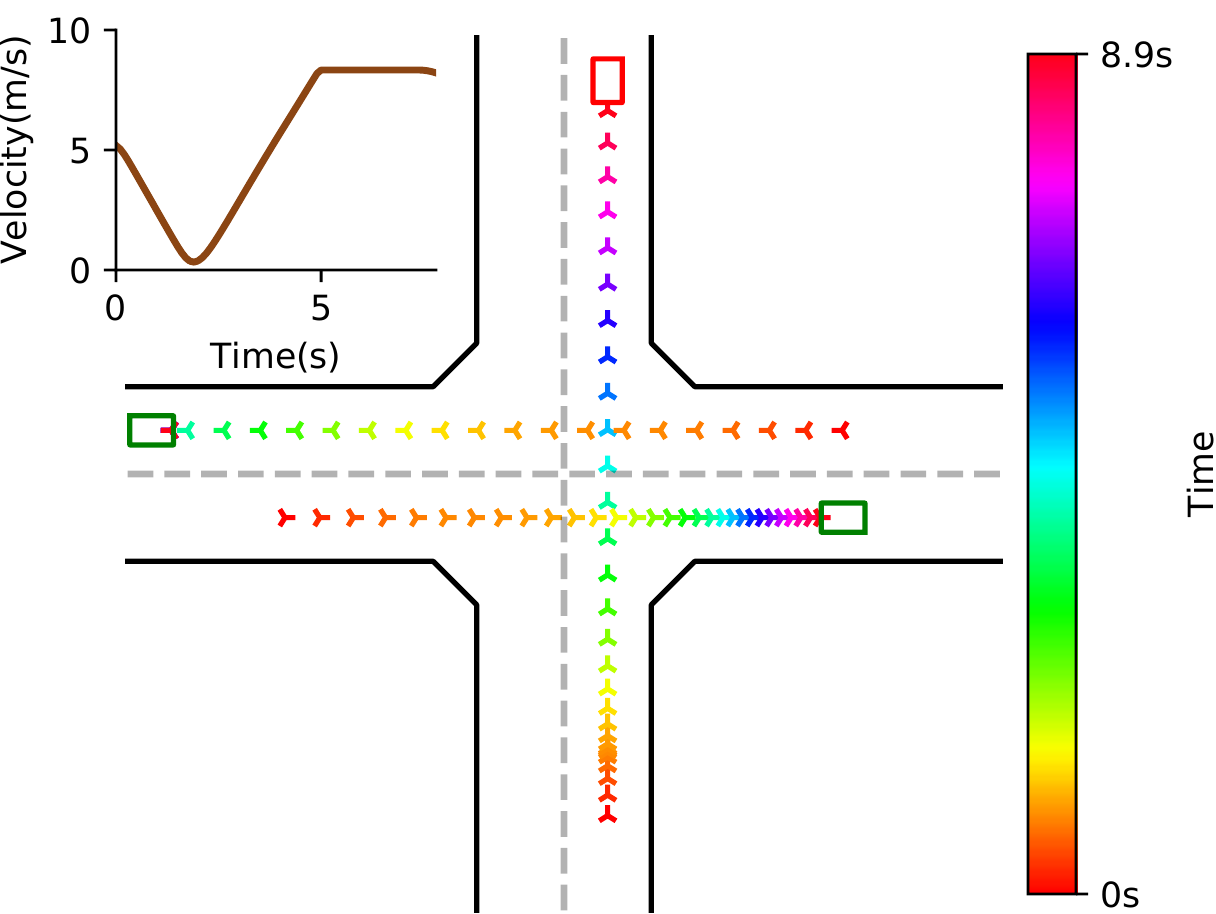}
  \caption{}
  \label{fig:HalfCheetah_ave}
\end{subfigure}
}
\caption{Result visualization of different policies. The brown line in each subplot shows the velocity of protagonist vehicle and the colorbar shows the location of all vehicles at each time step. Performance of  
(a) Minimax DSAC under aggressive mode (crossing in 8.5s).
(b) Minimax DSAC under conservative mode (crossing in 7.4s).
(c) Minimax DSAC under random mode (crossing in 6.1s).
(d) DSAC under aggressive mode (a failure pass, i.e., collision happens in 2.3s).
(e) DSAC under conservative mode (crossing in 7.6s).
(f) DSAC under random mode (crossing in 8.9s). 
}
\label{fig:visualization}
\end{figure*}
Fig.~\ref{fig:visualization} shows the control effect of trained policies for protagonist vehicle under the same behavior of adversarial vehicles. In aggressive mode, the protagonist vehicle of Minimax DSAC learned to decelerate priorly to wait until all adversary vehicles pass firstly, while the agent of DSAC suffers a collision resulting from its high speed. In conservative mode, both protagonist vehicles adopted similar strategy to pass successfully except that the Minimax DSAC gets a little less pass time. Under this environment, riding with high speed to pass firstly will encounter less collision and improve the pass efficiency. In random mode, our Minimax DSAC can adjust the speed more flexible to pass the intersection with a less pass time (6.1s) than standard DSAC (8.9s).

To sum up, although the Minimax DSAC obtained smaller average return during training process, it can maintain better performance when encountering different kinds of variations from environment.

\section{Conclusion}
In this paper, we combine the minimax formulation with the distributional framework to improve the generalization ability of RL algorithms, in which the protagonist agent must compete with the adversarial agent to learn how to behave well. Based on the DSAC algorithm, we propose the Minimax DSAC algorithm and implement it on the autonomous driving task at intersections. 
Results show that our algorithm significantly improves the protagonist agent's persistence to the variation from the environment.
This study provides a promising approach to accelerate the application of RL algorithms in real world like autonomous driving, where we always develop algorithms on the simulator which is distinct from the real environment.

\bibliographystyle{unsrt} %ieeetr国际电气电子工程师协会期刊
\bibliography{ref} % ref就是之前建立的ref.bib文件的前缀

\end{document}